%% file: main.tex
\pdfoutput=1

\documentclass[11pt]{article}

\usepackage{acl}

\usepackage{times}
\usepackage{latexsym}
\usepackage{booktabs}
\usepackage{microtype}
\usepackage{amssymb}
\usepackage{pifont}
\usepackage{xcolor} 
\usepackage{graphicx}
\usepackage{placeins}
\usepackage{natbib}
\usepackage{tablefootnote}
\usepackage{longtable}
\usepackage{cleveref}
\usepackage{enumitem}
\usepackage{tikz}
\usepackage{booktabs}
\usepackage{forest}
\usetikzlibrary{trees}
\usepackage{pdfpages}
\usepackage[T1]{fontenc}

\usepackage{tabularx}
\usepackage{multirow}
\usepackage{xspace}

\usepackage[textsize=footnotesize]{todonotes}
\usepackage{linguex}
\usepackage{subcaption}
\alignSubExtrue

\usepackage[verbose]{newunicodechar}

\usepackage{inconsolata}

\title{SemRel2024: A Collection of Semantic Textual Relatedness Datasets\\ for 13 Languages}

\author{Nedjma Ousidhoum$^{1}$\thanks{$^{*}$Equal contribution from first and second authors, authors 3 to 26 are alphabetically ordered.} , Shamsuddeen Hassan Muhammad$^{2*}$, Mohamed Abdalla, \\
{\bf Idris Abdulmumin$^3$, Ibrahim Said Ahmad$^{4}$, Sanchit Ahuja$^{5}$, Alham Fikri Aji$^6$,
} \\
{\bf Vladimir Araujo$^{7}$, Abinew Ali Ayele$^{8,9}$, Pavan Baswani$^{10}$, Meriem Beloucif$^{11}$,
}\\
{\bf Chris Biemann$^{8}$, Sofia Bourhim, Christine De Kock$^{12}$, Genet Shanko Dekebo$^{13}$,
}\\
{\bf Oumaima Hourrane, Gopichand Kanumolu$^{10}$, Lokesh Madasu$^{10}$, Samuel Rutunda$^{14}$,} \\
{\bf  Manish Shrivastava$^{10}$, Thamar Solorio$^{6}$, Nirmal Surange$^{10}$, Hailegnaw Getaneh Tilaye$^{15}$,}\\
{\bf Krishnapriya Vishnubhotla$^{16}$, Genta Winata$^{17}$, Seid Muhie Yimam$^{8}$, Saif M. Mohammad$^{18}$}
\\
 \footnotesize {$^1$Cardiff University, $^{2}$Imperial College London, $^{3}$Data Science for Social Impact Research Group, 
 University of Pretoria,} \\
 \footnotesize {$^4$Institute For Experiential AI,
 Northeastern University, $^5$BITS Pilani, $^6$ MBZUAI, $^{7}$KU Leuven,} \\
  \footnotesize {$^{8}$Universität Hamburg, Language Technology Group, $^{9}$Bahir Dar University, Faculty of Computing,$^{10}$IIIT Hyderabad,}\\
 \footnotesize {$^{11}$Uppsala University, $^{12}$The University of Melbourne, $^{13}$Adama Science and Technology University, $^{14}$Digital Umuganda,} \\
  \footnotesize {$^{15}$Kotebe University of Education, $^{16}$University of Toronto, $^{17}$HKUST, $^{18}$National Research Council Canada} \\
 \footnotesize \texttt{Contact: OusidhoumN@cardiff.ac.uk}
 }

\begin{document}
\maketitle
\begin{abstract}

Exploring and quantifying semantic relatedness is central to representing language and holds significant implications across various NLP tasks.
While earlier NLP research primarily focused on semantic similarity, often within the English language context, we instead investigate the broader phenomenon of semantic relatedness. In this paper, we present \textit{SemRel}, a new semantic relatedness dataset collection annotated by native speakers across 13 languages: \textit{Afrikaans, Algerian Arabic, Amharic, English, Hausa, Hindi, Indonesian, Kinyarwanda, Marathi, Moroccan Arabic, Modern Standard Arabic, Spanish,} and \textit{Telugu}. These languages originate from five distinct language families and are predominantly spoken in Africa and Asia -- regions characterised by a relatively limited availability of NLP resources. Each instance in the SemRel datasets is a sentence pair associated with a score that represents the degree of semantic textual relatedness between the two sentences. The scores are obtained using a comparative annotation framework. We describe the data collection and annotation processes, challenges when building the datasets, baseline experiments, and their impact and utility in NLP. 

\end{abstract}

\section{Introduction}

\begin{figure}[!ht]
    \centering
    \includegraphics[scale=0.45]{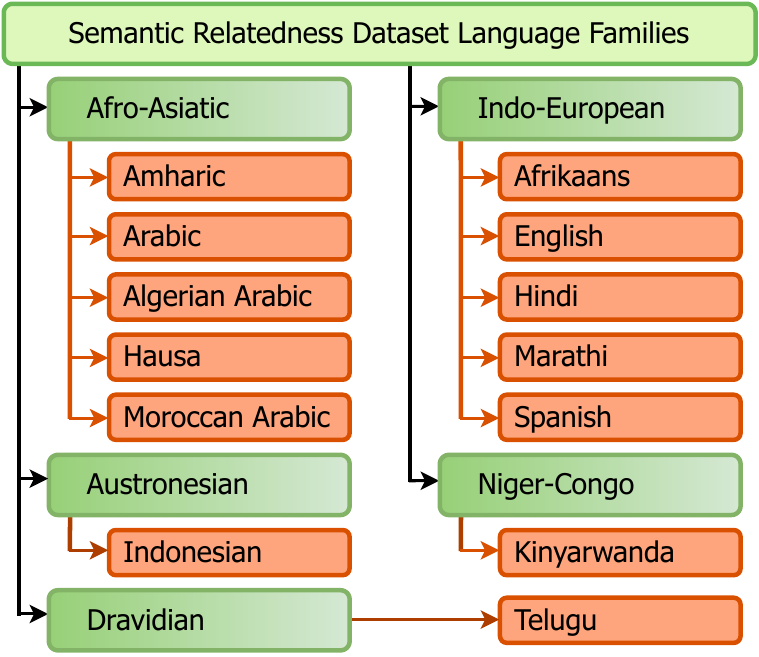}
    \caption{SemRel2024 languages and language families.}
    \label{fig:language_families}
    \vspace*{-5mm}
\end{figure}

Characterising the relationship between two units of text is an important component of constructing text representations. Within this context, semantic textual relatedness (STR) aims to capture the degree to which two linguistic units (e.g., words or sentences, etc.) are close in meaning \cite{mohammad2008measuring,mohammad2012distributional}. Two units may be related in a variety of different ways (e.g., by expressing the same view, originating from the same time period, elaborating on each other, etc.). On the other hand, semantic textual similarity (STS) considers only a narrow view of the relationship that may exist between texts (such as equivalence or paraphrase) which does not incorporate other dimensions of relatedness such as entailment, topic or view similarity, or temporal relations \cite{abdalla-etal-2023-makes,agirre-etal-2013-sem}. For example, \textit{`I caught a cold.'} and \textit{`I hope you feel better soon.'} would receive a low similarity score, despite the two being very related. In this work, we investigate the broader concept of semantic textual relatedness. 

STR is central to understanding meaning in text \cite{halliday1976cohesion,miller1991contextual,morris1991lexical} and its automation can benefit various downstream tasks such as evaluating sentence representation methods, question answering, and summarisation \cite{abdalla-etal-2023-makes,wang-etal-2022-just}.

\FloatBarrier
\begin{figure*}[!ht]
    \centering
        \includegraphics[clip, trim=2.6cm 6.5cm 2.6cm 6cm, width=0.8\textwidth]{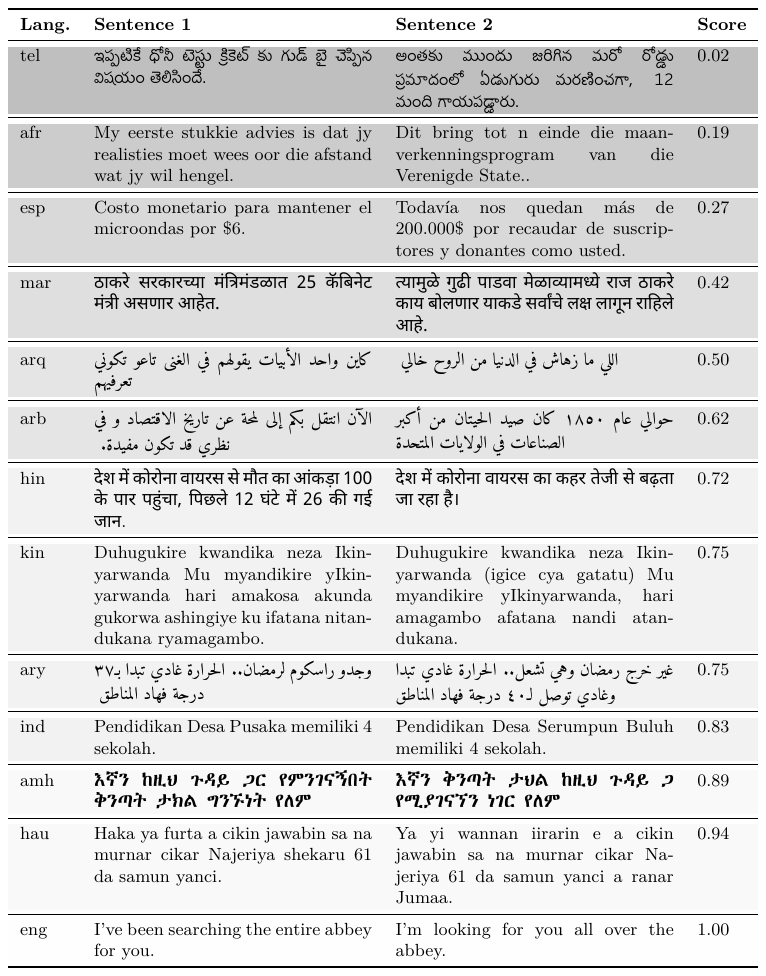}
    \vspace*{-3mm}
     \captionof{table}[]{Examples of sentence pairs and their corresponding scores (from 0 to 1) in the various SemRel2024 languages. Examples are sorted by score and rows with higher degrees of relatedness are lighter colored. The translations can be found in the Appendix.}
    \label{fig:SemRel_examples}
    \vspace*{-4mm}
\end{figure*}

Prior NLP work has mainly focused on textual similarity, largely due to a dearth of relatedness datasets. Of the existing STR and STS datasets, most are in the English. The few STR and STS resources which exist for non-high resource languages are composed of word-level or phrase-level pairings.
In this work, we curate 13 new monolingual STR datasets\footnote{Our team also created data for Punjabi. However, the sentence-pair selection procedure for it was markedly different than what we used for other languages. Therefore we do not include it here.} for Afrikaans \texttt{(afr)}, Amharic \texttt{(amh)}, Modern Standard Arabic \texttt{(arb)}, Algerian Arabic \texttt{(arq)}, Moroccan Arabic \texttt{(ary)}, English \texttt{(eng)}, Spanish \texttt{(esp)}, Hausa \texttt{(hau)}, Hindi \texttt{(hin)}, Indonesian \texttt{(ind)}, Kinyarwanda \texttt{(kin)}, Marathi \texttt{(mar)}, and Telugu \texttt{(tel)}. 

The datasets are composed of sentence pairs, each assigned a relatedness score between 0 (completely unrelated) and 1 (maximally related). With the aim of curating diverse STR datasets, the pairs of sentences were first selected from pre-existing datasets covering various topics and formality levels, e.g., news data, Wikipedia, and conversational data. Additionally, we selected pairs with a large range of expected relatedness values by considering lexical overlap, contiguity, topic coverage, and random pairings. To generate the relatedness scores, the sentence pairs were then annotated by native speakers who performed comparisons between different pairs of sentences using Best-Worst Scaling (BWS) \cite{louviere1991best}. BWS is known to avoid common limitations of traditional rating scale annotation methods \cite{kiritchenko-mohammad-2016-capturing,kiritchenko2017best}. The annotation process led to the high reliability of the final relatedness rankings in the different SemRel datasets. Our main contributions are as follows:\\[-22pt]
\begin{enumerate}
    \itemsep0em
     \item We present the first benchmark on semantic distance (similarity or relatedness) that includes low-resource African and Asian languages from five different language families (see Figure \ref{fig:language_families}). Although Africa and Asia are home to over 5,000 languages from over 20 language families and have the highest linguistic diversity, there is little publicly available data on these languages.
     \item We discuss general and language-specific challenges related to the data collection and annotation of the SemRel datasets.
     \item We present baseline experiments conducted in different monolingual and crosslingual settings to demonstrate the usefulness and potential of our dataset collection.
     \vspace*{-1mm}
 
\end{enumerate}

\input{tables/data_table}

To promote research in the field of semantic relatedness, we publicly released the SemRel2024 datasets as part of a shared task  that attracted a large number of participants interested in low-resource languages.\footnote{See \url{https://semantic-textual-relatedness.github.io} for more details.}

\section{Related Work}
\label{sec:related}

 The field of semantic relatedness in natural language processing covers a variety of approaches and techniques designed to measure the closeness in meaning between units, specifically words \cite{miller-1994-wordnet}, or sentences \cite{abdalla-etal-2023-makes}. 
 
 Most prior work focuses on STS, a narrower subset of STR, and often only covers high-resource languages such as English \cite{agirre2012semeval,agirre2013sem,agirre2014semeval,agirre2015semeval,agirre2016semeval,marelli2014sick}, Arabic, German, Spanish, Turkish \cite{ahmed2020multilingual, cer-etal-2017-semeval}, and Italian \cite{glavavs2018resource} with the only exception being Finnish, Slovene, Croatian \cite{glavavs2018resource,armendariz-etal-2020-semeval} and Farsi \cite{vulic2020multi}. 
 To overcome the scarcity of available resources, \citet{tang2018improving} proposed sentence-level, encoder-based methods leveraging English data to create Arabic, Spanish, Thai, and Indonesian datasets, whereas \citet{pandit2019improving} use traditional data augmentation methods to create Bangla data.
 
By comparison, this work is focused on the creation of resources for sentence-level STR in multiple low-resource languages. Here, the few works which exist for non-high-resource languages are at the word level (e.g., \citealp{yum2021word} for Korean). To our knowledge, the only corpora specially designed for semantic textual relatedness between pairs of sentences was created by \citet{abdalla-etal-2023-makes} for English. \citet{abdalla-etal-2023-makes} curated a dataset of 5,500 English sentence pairs annotated using a comparative annotation framework. Their dataset has since been used to evaluate embedding approaches \cite{wang2022relational} and other methods \cite{wang-etal-2022-just}.
The core of \citet{abdalla-etal-2023-makes} approach serves as the model for data annotations in this project. However, our work additionally explores new ways of data collection--curation, and several challenges had to be addressed when working with less-resourced languages.

\section{STR Data}\label{sec:data}

\subsection{Data collection}
\label{subsec:collection}
\begin{figure}[!ht]
    \centering
    \includegraphics[scale=0.48]{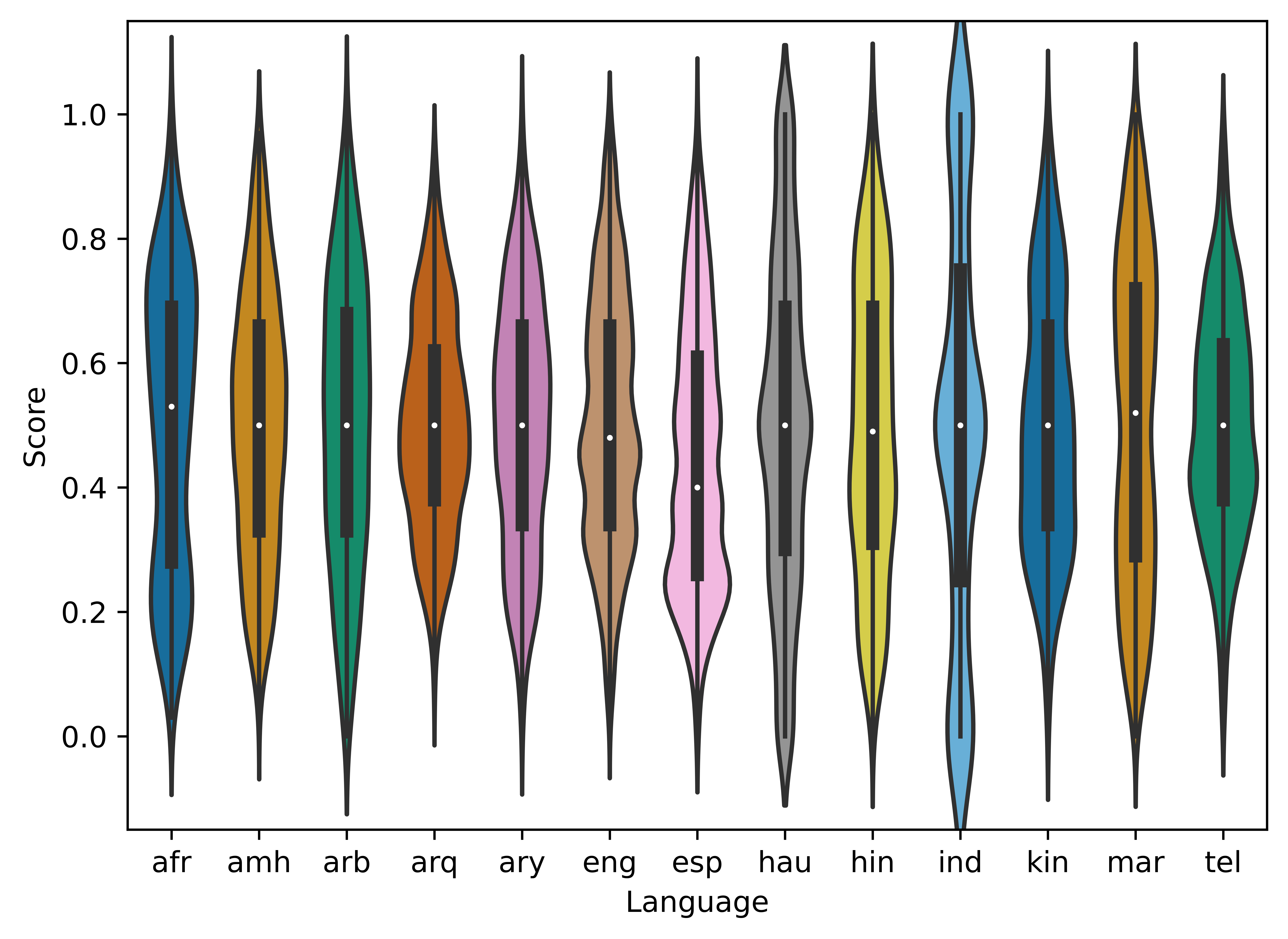}
    \caption{Violin plots representing the distributions of the relatedness scores. 
    For instance, the distribution of the Arabic (\texttt{arb}) dataset is unimodal, Marathi's (\texttt{mar}) is bimodal, and the Indonesian (\texttt{ind}) dataset's is trimodal.}
    \label{fig:relatedness_scores_dist}
    \vspace*{-5mm}
\end{figure}

A key step in the data creation process was identifying sources of text for each language and selecting sentence pairs. This was particularly challenging for low-resource languages such as Hausa, Kinyarwanda, and Algerian Arabic. Since arbitrarily selecting sentences and pairing them would lead to many unrelated instances, we relied on several heuristics, discussed in Section \ref{subsec:data_creation}, to ensure a wide range of scores for each language. Since these methods are highly corpus- and language-specific, the approaches used per language were determined by native speakers. We provide the data origin and the pairing approaches used for each language in Section \ref{sec:dataset_curation}. The composition of the resulting dataset is summarised in Table \ref{tab:data_collection} and the distribution of the relatedness scores across the datasets are illustrated in Figure \ref{fig:relatedness_scores_dist}. 

\subsubsection{Sentence pairing heuristics} \label{subsec:data_creation}

Given a set of texts in a target language, careful consideration was given to the construction of sentence pairs to ensure that the pairs would exhibit relatedness scores varying from completely unrelated to very related. Since random selection would result in many unrelated pairs, we paired sentences mainly based on five methods previously defined by \citet{abdalla-etal-2023-makes} (described below). In cases where the pairs produced by the five methods were qualitatively judged to be insufficiently varied, we manually selected some instances to balance the data so that we have sufficient number of instances for each band of relatedness (high, medium, low, or unrelated).

\paragraph{Lexical overlap} Pairs are selected with various amounts of lexical overlap. That is, one or more words/tokens in common, with or without using TF-IDF normalisation. This method is expected to produce a wide range of relatedness values, and was used in most low-resource languages. 

\paragraph{Contiguity/Entailment} We select pairs of sentences that appear one after the other in a paragraph or a social media thread. This method is likely to produce pairs of sentences that are somewhat related and can contribute to representing the low to medium ranges of relatedness.

\paragraph{Paraphrases or Machine Translation (MT) paraphrases} This method consists of selecting pairs of sentences from paraphrase or MT data. For MT, we pivot across the translation and back to the source language to generate a new sentence and pair it with the original. However, many low-resource Asian and African languages lack reliable MT resources.

\paragraph{Semantically similar instances} Semantically similar sentences are selected from a publicly available dataset such as the STS dataset by \citet{cer2017semeval} or manually identified by a native speaker in order to include highly related instances and balance the dataset.

\paragraph{Random selection} Random sentences are selected. This method is expected to represent the low to medium ranges of relatedness.

 \paragraph{Manual check} In cases where the pairs produced by the above methods were qualitatively judged to be insufficiently varied, instances were manually selected to balance the data so that there were sufficient number of instances for each band of relatedness (high, medium, low, or unrelated).
 This can apply to any range of relatedness (i.e., high, medium, low, or unrelated).
\input{tables/SHR}

\subsubsection{Data curation}\label{sec:dataset_curation}

Since most of the SemRel languages are low-resource, the domain, (in)formality, and diversity of the sentence pairs were highly dependent on the publicly available corpora.
We aimed to collect datasets with average-length sentences, free of offensive utterances, and as diverse as possible. As such, data instances were extracted for each language using a tailored combination of the heuristics described in Section~\ref{subsec:data_creation}. We used further pre-processing, post-processing, and data analysis methods (discussed below) to avoid incoherence and unnaturalness.

\paragraph{English and Spanish}
As English and Spanish are high-resource languages, we sampled sentences from various sources to capture a wide variety of sentence structure, formality, and grammaticality in texts. As shown in Table \ref{tab:data_collection}, we paired sentences in a number of ways that include lexical overlap, entailment, similarity and paraphrases. The English dataset includes sentences that have the same meaning but a different formality collected from the Formality dataset \cite{rao-tetreault-2018-dear}, tweets \cite{mohammad2017stance}, paraphrases from machine translation systems extracted from the ParaNMT dataset \cite{wieting2017paranmt}, book reviews from Goodreads \cite{wan2018item}, pairs of premises and hypotheses from the SNLI dataset \cite{bowman-etal-2015-large}, and semantically similar sentences \cite{cer2017semeval}.

Similarly, we select pairs of Spanish sentences from semantic similarity datasets such as STS \cite{agirre2014semeval,agirre2015semeval,cer2017semeval}, entailment datasets such as SICK-es \cite{huertastato2021silt} and NLI-es \cite{araujo-etal-2022-evaluation}, and paraphrasing datasets such as PAWS-X \cite{yang-etal-2019-paws}.
We also sampled contiguous sentences from XL-Sum \cite{hasan2021xl} and BSO DiscoEval Spanish \cite{araujo-etal-2022-evaluation}, and we included questions of different types from Spanish QC \cite{a-garcia-cumbreras-etal-2006-bruja}.

\paragraph{Arabic Variations: Modern Standard Arabic, Algerian, and Moroccan Arabic}
Arabic is known for diglossia \cite{ferguson1959diglossia}, meaning that Arabic varieties are used for different contexts. For instance, Modern Standard Arabic is usually used in formal and academic communication while dialects are typical for conversational settings. The various sources of the Arabic data are somewhat reflective of the distinct language usage scenarios.

Therefore, for Modern Standard Arabic (MSA), we used two datasets from different domains: TED Talk subtitles \cite{zong-2015-improving} on science, society, and art and news articles on economics \cite{arabic_news_data}. In addition to sentences with lexical overlap, we selected contiguous sentences in Ted Talk subtitles to include different degrees of relatedness, and as some sentences in the subtitles were slightly ungrammatical, we corrected them based on the standard Arabic grammar rules.
For Algerian Arabic, we used CalYou \cite{abidi2017calyou}, a dataset composed of YouTube comments collected from major Algerian YouTube channels by 2017, and the Algerian instances spoken in two major Algerian towns (Algiers and Annaba) present in PADIC: Parallel Arabic Dialect Corpus \cite{meftouh2015machine}. We used lexical overlap to pair sentences, picked contiguous ones in a conversation in PADIC, and added randomly or manually selected sentence pairs to balance the relatedness score distribution in the dataset. For both MSA and Algerian Arabic, we allowed short sentences as Arabic is highly inflectional.
For Moroccan Arabic, we used headlines from the Goud.ma dataset introduced by \citet{issam2022goudma} and the Moroccan Arabic sentences were paired based on lexical overlap.

\paragraph{Afrikaans} 
The Oscar dataset \cite{ortiz-suarez-etal-2020-monolingual} was used as basis for the Afrikaans corpus. We chose sentences from news articles, blogs, reviews, and recipes. We also excluded sentences from religious texts and academic articles after observing that these did not produce high-quality pairs. We further excluded a number of advertorial texts that appear to be low-quality translations. All instances were then manually assessed for grammar and ungrammatical sentences were discarded. 
Sentences were paired if they had an overlap of at least five tokens and at least three non-overlapping tokens with matches within the same article only. Random sentence pairs were also included to calibrate the dataset.

\paragraph{Amharic, Hausa, Kinyarwanda}

For Amharic, we paired sentences present in news articles from different Ethiopian news outlets \cite{fi13110275}. Similarly, the Hausa and Kinyarwanda datasets include pairs of sentences from news articles collected by \citet{Abdulmumin_2019} and \citet{niyongabo-etal-2020-kinnews}, respectively. Sentences shorter than five words and longer than 20 were excluded, and pairs were created using lexical overlap. Additionally, for Amharic, we excluded sentences with mixed languages to avoid confusing the annotators.

\paragraph{Indonesian}
For Indonesian, we collected sentences from Wikipedia texts present in the ROOTS split \cite{laurenccon2022bigscience,setya2018semi} and the IndoSum~\cite{kurniawan2018indosum} datasets. IndoSum is a human-written summarization dataset consisting of pairs of news articles with abstractive summaries. We parsed both corpora at a sentence level and only selected sentences that were composed of five to fifteen words.

\paragraph{Hindi, Marathi, and Telugu}
As these languages lack publicly available resources, especially Marathi and Telugu, we used the Mukhyansh dataset only \cite{madasu-etal-2023-mukhyansh} to create the sentence pairs. It is composed of news headlines and their corresponding articles and is diverse in nature. We created instances using lexical overlap, paraphrase generation, contiguous sentence selection, and random sentence selection to balance the data.

\subsection{Data annotation and challenges}
\input{tables/data_splits}

\paragraph{Annotation process}
Similarly to \citet{abdalla-etal-2023-makes}, we used BWS to annotate our data instances and generate an ordinal ranking of instances\footnote{The tuples were generated using the BWS scripts provided by \cite{kiritchenko2017best}: \url{http:// saifmohammad.com/WebPages/BestWorst.html}.}. 
Although pairwise comparisons are more reliable than simply labelling the sentence pairs as related or unrelated, it is a time-consuming process if performed on a large dataset as it requires $N \times N = N^2$ comparisons if performed on a dataset of $N$ instances. Best-worst scaling mitigates this issue according to \citet{kiritchenko2017best} as it leads to reliable scores from about $2 \times N$ comparisons of 4-instance tuples. 

BWS requires fewer labels \cite{louviere1991best},
in our case, given four instances (i.e., pairs of sentences) $p_i$ with $0 \leq i < 4$, for a tuple: $\langle p_0,p_1,p_2,p_3 \rangle$,  if $p_0$ is marked as most related and $p_3$ as least related, then we know that $p_0>p_1$, $p_0>p_2$, $p_0>p_3$, $p_1>p_3$, and $p_2>p_3$ ($<$ and $>$ refer to less and more related, respectively). 
We then use these inequalities to compute real-valued scores that consist of the fraction of times a pair $p_i$ was chosen as the most related minus the fraction of times $p_i$ was chosen as the least related. Then, an ordinal ranking of sentence pairs is generated \cite{Orme2009MaxDiffA,flynn2014best}.

Furthermore, the notions of \textit{related} and \textit{unrelated} have fuzzy boundaries with no singular accepted definition in the literature. Different people and different language cultures may have several intuitions of where such a boundary exists. Therefore, by using comparative annotations and relying on the intuitions of fluent speakers for each language to choose between sentence pairs, we can avoid ill-defined categories. This is in line with our goal of capturing common perceptions of semantic relatedness (i.e., what is believed by the vast majority) instead of ``correct'' or ``right'' rankings.

\paragraph{Instructions}
We selected native speakers to annotate the sentence pairs. Then, given a set of four sentence pairs, annotators were tasked with reporting on their relative relatedness. Concretely, given 4 sentence pairs, each of the form \textit{[sentence A, sentence B]}, the task was to select the sentence pair that is the \textit{most related} (i.e., sentence A is closest in meaning to sentence B) and the sentence pair that is the \textit{least related} (i.e., sentence A is farthest in meaning to sentence B). The full instructions can be found in the Appendix.

In the guidelines, it was noted that sentence pairs that are more specific in what they share tend to be more related than sentence pairs that are only loosely about the same topic. Furthermore, if one or both sentences have more than one interpretation, the annotators have to consider the closest meanings. 

Overall, by manually examining the annotations, we noted that the BWS framework does lead to more robust annotations. However, a downside is the fact that annotating one instance could take more than one minute and the task can be challenging since many instances to be compared can be similarly (un)related.

\paragraph{Annotation reliability}
 In Table \ref{tab:shr_scores}, we report the number of annotators and the split-half reliability (SHR) \cite{cronbach1951coefficient,kuder1937theory} scores for each of the datasets. 
 SHR measures the degree to which repeating the annotations results in similar relative rankings of the instances. First, it splits the 4-tuple annotations into two bins. Then, the annotations for each bin are used to generate two different independent relatedness scores, and the Spearman correlation between the two sets of scores is calculated to estimate the closeness of the two rankings.
 A high correlation indicates that the annotations are reliable. This process is repeated 1,000 times and the correlation scores are averaged similarly to \citet{abdalla-etal-2023-makes}.
 Overall the scores in Table \ref{tab:shr_scores} vary between 0.64 and 0.96, which indicates a high annotation reliability.

\paragraph{Disagreements}
We inspected annotators with large disagreements to ensure the annotation procedure was correctly followed (i.e., their annotations made sense for native speakers). Very strong disagreements would serve as a red flag of poor data quality resulting in a more thorough review of the annotation quality. Hence, as a sanity check, we examined whether sentences with high relatedness scores were more semantically related than those with low relatedness scores. The specific procedure for ensuring data quality depends on the annotation procedure of the team (e.g., those using AMT vs those who did not).
Note that disagreements were not deleted, as they can serve as a useful signal for BWS. That is, for a tuple $\langle p_0,p_1,p_2,p_3 \rangle$, when annotators disagree on what is most related (e.g., $p_1$ or $p_3$), then it is an indication that $p_1$ and $p_3$ may be semantically close to each other. As all tuple annotations (twice the number of instances) are used to determine the final scores of the sentence pairs, this disagreement would lead to the two pairs ($p_1$ and $p_3$) getting scores that are close to each other. On the other hand, if a sentence pair consistently occurs in 4-tuples that have very low annotator agreement, then it is likely that the sentence pair is the source of disagreement. This can be due to various reasons such as the language use, code-switching, or the annotator's familiarity with the topics discussed.

Besides sharing our datasets with the community, we also make the full 4-tuple annotations public.

\subsection{Postprocessing and data quality control}
\paragraph{Quality control}

 For our final dataset, we carried a data post-processing step to ensure that:
 \begin{itemize}[noitemsep,nolistsep] 
     \item no instances are repeated;
     \item the data does not include invisible characters, incorrectly rendered emoticons, or garbled encoding characters;
    \item texts are fully anonymised (deleting emails and IDs if they occur, replacing @mentions with @<username>, and replacing any URLs with non-identifiable placeholders);
    \item the data does not include a high amount of expletives or inappropriate language; and
    \item the data is balanced. 
 \end{itemize}
 
\paragraph{Manual Spotchecks}

Finally, a team of native speakers manually spot-checked the scores to make sure that the relatedness scores made sense and to supplement the quantitative evaluation based on SHR.

\section{Experiments}
\subsection{Data}
We use the splits reported in Table \ref{tab:data_splits}. For the languages without training data (afr, arb, hin, ind), we only report experiments in unsupervised and crosslingual settings. For English, we use the STR-2022 dataset \cite{abdalla-etal-2023-makes} for training and we use our newly created dataset for testing.

\subsection{Experimental setup}

\input{tables/updated_table_no_pan}
We report the Spearman correlation scores between the predicted labels and the gold standard ones for the different languages in three main settings:
\begin{itemize}[noitemsep,nolistsep] 
\item \textbf{Supervised} systems trained on the labeled training datasets provided.
\item \textbf{Unsupervised} systems developed without the use of labeled datasets pertaining to semantic relatedness or semantic similarity between units of text of more than two words long in any language. 
\item \textbf{Crosslingual} systems developed without the use of any labeled semantic similarity or semantic relatedness datasets in the target language and with the use of data from at least one other language included in SemRel.
\end{itemize}

In our experiments, we use:
\begin{itemize}[noitemsep,nolistsep] 
     \item a simple baseline based on the number of shared words (lexical overlap),
    \item sentence embeddings (LaBSE \cite{feng2020language}, SentenceBERT \cite{reimers-2019-sentence-bert}), and
    \item standard encoder-based embeddings. 
\end{itemize}
\subsection{Lexical Overlap}

As shown in \Cref{tab:encoder-based-models}, we report a simple lexical overlap baseline which consists of the Dice coefficient between two sentences A and B: the number of unique unigrams occurring in both sentences, adjusted by their lengths \cite{abdalla-etal-2023-makes}: 
\begin{equation}
        \frac{2 \times | unigram(A) \cap unigram(B) |}{ | unigram(A) + unigram(B) |}
\end{equation}

\subsection{Supervised and Crosslingual settings}
We use LaBSE (Label Agnostic BERT Sentence Embeddings) \cite{feng2020language} which can map 109 languages into a shared vector space. With the embeddings covering all the SemRel languages, we report baseline results using the default hyperparameters set in the sentence-transformers repository\footnote{\url{https://github.com/UKPLab/used-transformers}}. Our experiments are conducted:
\begin{itemize}[noitemsep,nolistsep] 
    \item using the predefined setup without further fine-tuning,
    \item by fine-tuning the LaBSE model on our training data using a cosine similarity loss.
\end{itemize}
 We report the scores on the test sets in both setups in 
 Appendix A, Table \ref{tab:sbert_baselines}.

For the crosslingual baselines, we fine-tune LaBSE on the English training set and test on all the other datasets except English. On the other hand, when testing on the English dataset, we use the Spanish training set to fine-tune LaBSE.

\subsection{Unsupervised settings}
We used the standard encoder-based monolingual and multilingual language models on our datasets\footnote{We use the standard models from HuggingFace \cite{wolf-etal-2020-transformers}.}. We experiment with:
\begin{itemize}[noitemsep,nolistsep]
    \item multilingual BERT (mBERT) \cite{devlin-etal-2019-bert}, XLMRoberta (XLMR) \cite{conneau-etal-2020-unsupervised} for all 13 languages,
    \item monolingual models:
        \begin{itemize}[noitemsep,nolistsep]
            \item AfroXLMR \cite{alabi-etal-2022-adapting} for Afrikaans, Amharic, Hausa, Kinyarwanda, 
            \item Indic-BERT \cite{kakwani-etal-2020-indicnlpsuite} for Hindi, Marathi, and Telugu, 
            \item BERT \cite{devlin-etal-2019-bert} for English, 
            \item MARBERT, ARBERT \cite{abdul-mageed-etal-2021-arbert} and Arabic BERT \cite{safaya-etal-2020-kuisail} for Arabic, 
            \item BETO \cite{CaneteCFP2020}, ALBETO \cite{canete-etal-2022-albeto}, and RoBERTa-BNE \cite{maria} for Spanish, 
            \item Amharic RoBERTa (AmRoBERTa) \cite{fi13110275} for Amharic, 
            \item DziriBERT \cite{dziribert} for Algerian Arabic,
            \item RoBERTa based model (HauRoBERTa) for Hausa \cite{adelani-etal-2022-masakhaner}.
        \end{itemize} 
\end{itemize}
We report the Spearman correlation scores with cosine similarity scores for all the BERT-based models in \Cref{tab:encoder-based-models}. Additional results using BERTScore \cite{bert-score} for mBERT and XLMR are shared in the Appendix 
(see \Cref{tab:other_baselines}).

\subsection{Experimental results}

Table \ref{tab:encoder-based-models} shows the Spearman correlation scores for the three setups: supervised, unsupervised, and crosslingual for all thirteen languages. For the unsupervised models, we report the results using all pretrained models including mBERT \cite{devlin-etal-2019-bert} and XLMR \cite{conneau-etal-2020-unsupervised}. Additionally, we report the Spearman correlation for experiments with monolingual language-specific models for each language -- models that have been trained in these specific languages. 

For the general setup, we note that, except for Amharic and Kinyarwanda, mBERT outperforms XLMR in all languages by a significant margin. For Amharic, mBERT's correlation score with the gold labels is 0.13, whereas XLMR's is three times higher with a 0.57 correlation score.  Surprisingly, even though Arabic is a high-resource language, the Spearman correlation score is relatively low in comparison to all the high-resource languages, with Spanish achieving the best results. This could be due to the size of the Arabic data being smaller. 

For the language-specific models, the results are highly tied to the language. In cases such as Amharic for example, AmRoBERTa significantly improves the score by 0.27 points, whereas AfroXLMR hurts the performance for all African languages. 

Similarly to the unsupervised setup, high-resource languages have the highest scores in supervised and crosslingual settings. Overall, we report relatively higher correlation scores which vary between 0.40 and 0.88.

\section{Conclusion}
We presented SemRel, a new collection of semantic textual relatedness datasets in 13 languages with the majority predominantly spoken in Africa and Asia and considered low-resource. The sentence pairs contained in the datasets are annotated by native speakers and are associated with fine-grained relatedness scores. We reported the details related to the data curation and emphasised the challenges faced when dealing with low-resource languages.

We publicly release the datasets as well as other resources, such as the annotation guidelines and full labels for the research community interested in semantic relatedness, low-resource languages, and disagreements.

\section{Limitations}
We acknowledge that there is no formal definition of what constitutes semantic relatedness. Hence, the annotations may be subjective. To mitigate the issue we share our guidelines and annotated instances so researchers in the community can expand on our work, replicate, and study the disagreements in our data. We are also aware of the limited number of data sources and data variety in some low-resource languages involved. We do not claim that the datasets released represent all variations of these languages but they remain a good starting point as they were carefully picked, labelled, and processed by native speakers.

Although our collection is comprised of multiple datasets, the size of the data is limited, thus it cannot be the only source used for tasks that require a large amount of data such as language identification.

\section{Ethical Considerations}

All the annotators involved in this study were either volunteers or were paid more than the minimum wage per hour and any demographic information reported in the Appendix was shared with consent. The data that was further annotated was publicly available and is cited in our paper.

Similarly to \citet{abdalla-etal-2023-makes}, we acknowledge all the possible socio-cultural biases that can come with our data, due to the data sources or the annotation process. When building our datasets, we did avoid instances with inappropriate or offensive utterances but we might have missed some.
Our goal was to identify common perceptions of semantic relatedness by native speakers and our labels are not meant to be standardised for any given language. Note that we build datasets for low-resource languages but we do not claim in any way that these are fully representative of their usage.

\section*{Acknowledgements}
We thank our annotators for labelling the data and for the insightful comments as well as Zara Siddique for providing additional insights.

Thanks to Dimosthenis Antypas, Joanne Boisson, Hsuvas Borkakoty for the helpful feedback.

\bibliography{anthology,custom}
\bibliographystyle{acl_natbib}

\appendix

\input{appendix}
\end{document}

%% file: tables/data_table.tex
\begin{table*}[h!]
\small
\centering
    \begin{tabular}{lll}
    \toprule[1.2pt]
        \textbf{Lang.} & \textbf{Curation technique} & \textbf{Data Sources} \\ \midrule
        \texttt{afr} & Overlap, Random selection, Manual check & News data, reviews, recipes, blogs.\\
        \texttt{amh} & Overlap, Similarity, Random selection, Manual check   & News data, crawling. \\
        \texttt{arb} & Overlap, Contiguity, Random selection, Manual check& Ted talk subtitles, news data. \\
        \texttt{arq} & Overlap, Contiguity, Random selection, Manual check & YouTube comments, conversational data.\\
        \texttt{ary} & Overlap &News data.\\
        \texttt{eng} & Overlap, Similarity, Paraphrases, Contiguity, Randomness& Book reviews, news data, tweets, other.\\
        \texttt{esp} & Overlap, Contiguity, Similarity & Movie reviews, news data, other.\\
        \texttt{hau} & Overlap& News data.\\
        \texttt{hin} & Overlap, Similarity, Contiguity, Paraphrase, Randomness &News data, other. \\
        \texttt{ind} & Overlap &Wikipedia, news data.\\
        \texttt{kin} & Overlap & News data. \\
        \texttt{mar} & Overlap, Similarity, Contiguity, Paraphrase, Randomness& News data, other.\\
        \texttt{tel} & Overlap, Similarity, Contiguity, Paraphrase, Randomness&News data, other.\\ \bottomrule
    \end{tabular}
\caption{The curation techniques used for data creation. We list the main textual sources present in the datasets we used for instance creation. More details are shared in Section \ref{sec:dataset_curation}.}
\label{tab:data_collection}
\vspace*{-1mm}
\end{table*}

%% file: tables/SHR.tex
\begin{table*}[]
    \centering
    \small
      \begin{tabular}{l|rrrrrrrrrrrrr}
         \toprule[1.2pt]
         \textbf{Language} & \texttt{afr} & \texttt{amh} & \texttt{arb} & \texttt{arq} & \texttt{ary} & \texttt{eng} & \texttt{esp} & \texttt{hau} & \texttt{hin} & \texttt{ind} & \texttt{kin} & \texttt{mar} & \texttt{tel}  \\ 
    \midrule
             \textbf{\#Ann/tuple} & 2 & 4 & 2-3 & 2 & 2 & 2-4 & 2-4 & 2-4 & 4 & 2 & 2 & 2-3 & 4 \\ 
        \textbf{SHR train/dev} & 0.85 & 0.90 & 0.86 & 0.64 & 0.77 & 0.84 & 0.70 & 0.74 & 0.93 & 0.68 & 0.74 & 0.92 & 0.79 \\ 
        \textbf{SHR test} & 0.85 & 0.90 & 0.86 & 0.64 & 0.77 & 0.80 & 0.70 & 0.74 & 0.94 & 0.68 & 0.74 & 0.96 & 0.96 \\

    \bottomrule
    \end{tabular}
    \vspace*{-1mm}
    \caption{SHR (split-half reliability) scores for each of the 
    dataset splits and numbers of unique annotations per tuple (\#Ann/tuple). As some languages (\texttt{eng}, \texttt{hin}, \texttt{mar}, and \texttt{tel}) had splits annotated 
    in separate annotation efforts (instead of one combined annotation), we report the SHR scores for both.}
    \label{tab:shr_scores}
    \vspace*{-2mm}
\end{table*}
\label{subsec:annotation}

%% file: tables/data_splits.tex
\begin{table*}[h]
    \centering
    \small
     \begin{tabular}{lrrrrrrrrrrrrr}
        \toprule[1.2pt]
         \textbf{Data}& \texttt{afr} & \texttt{amh} & \texttt{arb} & \texttt{arq} & \texttt{ary} & \texttt{eng} & \texttt{esp} & \texttt{hau} & \texttt{hin} & \texttt{ind} & \texttt{kin} & \texttt{mar} & \texttt{tel}  \\
        \midrule
        \textbf{Train} & - & 992 & - & 1,261 & 924 & 5,500 & 1,562 & 1,736 & - & - & 778 & 1,200 & 1,170 \\
        \textbf{Test} & 375 & 171 & 595 & 583 & 426 & 2,600 & 600 & 603 & 968 & 360 & 222 & 298 & 297 \\
        \textbf{Dev} & 375 & 95 & 32 & 97 & 71 & 250 & 140 & 212 & 288 & 144 & 102 & 293 & 130 \\
        \midrule
        \textbf{Total} & 700 & 1,258 & 627 & 1,941 & 1,421 & 8,350 & 2,302 & 2,551 & 1,256 & 504 & 1,102 & 1,791 & 1,597 \\
        \bottomrule
    \end{tabular}
    \caption{Number of instances in the training, dev, and test sets for the different datasets. The languages with no training data (\texttt{afr, arb, hin, ind}) were only used in unsupervised and cross-lingual settings.}
    \label{tab:data_splits}
    \vspace*{-3mm}
\end{table*}

%% file: tables/updated_table_no_pan.tex
\begin{table*}[!ht]
    \centering
\small
     \begin{tabular}{lccccccccccccc}
        \toprule[1.2pt]
         & \texttt{afr} & \texttt{amh} & \texttt{arb} & \texttt{arq} & \texttt{ary} & \texttt{eng} & \texttt{esp} & \texttt{hau} & \texttt{hin} & \texttt{ind} & \texttt{kin} & \texttt{mar} & \texttt{tel}  \\
        \midrule
        \textbf{Overlap}&0.71 & 0.63 & 0.32 & 0.40 & 0.63 & 0.67 & 0.67 & 0.31 & 0.53 & 0.55 & 0.33 & 0.62 & 0.70 \\        

        \midrule
        & \multicolumn{13}{c}{\textbf{Unsupervised (Multilingual)}} \\\cmidrule(lr){2-14}
        \textbf{mBERT} & 0.74 & 0.13 & 0.42 & 0.37 & 0.27 & 0.68 & 0.66 & 0.16 & 0.62 & 0.50 & 0.12 & 0.65 & 0.66 \\
        \textbf{XLMR} & 0.56 & 0.57 & 0.32 & 0.25 & 0.17 & 0.60 & 0.69 & 0.04 & 0.51 & 0.47 & 0.13 & 0.60 & 0.58 \\
        \midrule
        & \multicolumn{13}{c}{\textbf{Unsupervised (Monolingual)}} \\
        \cmidrule(lr){2-14}
        \textbf{AfroXLMR} & 0.45 & 0.40 & 0.18 & - & - & 0.30 & - & 0.07 & - & - & 0.16 & - & - \\
        \textbf{ALBETO} & - & - & - & - & - & - & 0.62 & - & - & - & - & - & - \\
        \textbf{AmRoBERTa} & - & 0.72 & - & - & - & - & - & - & - & - & - & - & - \\
        \textbf{ARBERT} & - & - & 0.56 & - & - & - & - & - & - & - & - & - & - \\
        \textbf{\texttt{arb} BERT} & - & - & 0.31 & - & - & - & - & - & - & - & - & - & - \\
        \textbf{BETO} & - & - & - & - & - & - & 0.68 & - & - & - & - & - & - \\

   \textbf{DziriBERT} & - & - & - & 0.43 & - & - & - & - & - & - & - &- & \\
     
           \textbf{Indic-BERT} & - & - & - & - & - & - & - & - &0.40 & - & - & 0.41\\
  
       \textbf{MARBERT} & - & - & 0.29 & - & - & - & - & - & - & - & - & - & - \\

        \textbf{RoBERTa-BNE} & - & - & - & - & - & - & 0.66 & - & - & - & - & -  & - \\
        \textbf{HauRoBERTa} & - & - & - & - & - & - & - & 0.12 & - & - & - & -  & - \\
        \midrule
        \textbf{} & \multicolumn{13}{c}{\textbf{Supervised}}  \\
        \midrule
         \textbf{LaBSE}& - & 0.85 & - & 0.60  & 0.77 & 0.83 & 0.70 & 0.69 & -  & - & 0.72 & 0.88 & 0.82 \\

        \midrule
        \textbf{} & \multicolumn{13}{c}{ \textbf{Crosslingual}} \\
        \midrule
        \textbf{LaBSE} & 0.79 & 0.84 & 0.61 & 0.46 & 0.40
 & 0.80 & 0.62 & 0.62 & 0.76 & 0.47 & 0.57 & 0.84 & 0.82 \\
        \bottomrule[1.2pt]
    \end{tabular}
    \vspace*{-1mm}
    \caption{Spearman correlation scores for different fine-tuned models in the three settings that we describe (supervised, unsupervised, and crosslingual) in addition to a simple lexical overlap baseline (Overlap).} 
    \label{tab:encoder-based-models}
    \vspace*{-3mm}
\end{table*}

%% file: appendix.tex
\section*{Appendix} \label{sec:appendix}

\section{Annotation}

\input{tables/data_sources}

\subsection{Pilot data annotation}
To assess the different pairing techniques and the potential annotation challenges, we run a pilot annotation task on 20 to 100 pairs of sentences for each language before proceeding with larger annotation batches. This helped us assess the difficulties related to the annotation task and the choices to be made for the final data processing step. For instance, if highly related and unrelated pairs were occurring too often in the tuples, we reduced the percentages of both highly related and unrelated pairs by changing or calibrating the data sources if possible, prioritising other pairing techniques, or including an extra preprocessing step (e.g., paraphrase detection).

\subsection{Data Pre-processing Tools}
We used NLTK tools for parsing Afrikaans and Indonesian in addition to manual verification. For instance, as for Indonesian, the NLTK sentence parses generated many errors due to common Indonesian abbreviations that involve \textit{'.'}, which the sentence parser mistakenly detects as the end of a sentence, we added new abbreviations for parsing ('ir.', 'kh.', 'h.', 'drs.', 'drg.', 'rm.', 'bp.', 'bpk.', 'tgl.', 'no.', 'jl.', and 'jln.')

\subsection{Information about the Annotators}
We report on the demographic information of the volunteers who agreed to share them.
\paragraph{Afrikaans}
Paid native speakers.
\paragraph{Amharic}
Paid Amharic native speakers, 3 women and 5 men from different social, cultural, and ethnic backgrounds (Amhara, Guragie, Wolyta, Sidama, and Oromo). 
\paragraph{Modern Standard Arabic and Algerian Arabic} Native speakers, 2 men, 2 women, university degree holders, ages vary between 23 to 56, paid above the minimum wage.
\paragraph{Moroccan Arabic}
Volunteer native speakers, 3 women, 1 man, university degree holders, volunteers.
\paragraph{English and Spanish}
Amazon Mechanical Turkers with high approval rates (98\% for English) paid above the US minimum wage.
\paragraph{Hausa}
Paid native speakers, 3 women, 1 man, age: 28 to 30, bachelor's degree holders.
\paragraph{Hindi and Marathi}
Paid native speakers.
\paragraph{Telugu}
Volunteer native speakers.

\subsection{Annotation Guidelines}
\textbf{You will be given four sentence pairs} (i.e., 4 pairs of the form [sentence A, sentence B]). Your task is to judge the relatedness of each pair (sentence A and sentence B) and \textbf{tell us}:
\begin{itemize}
    \item the sentence pair that is the \textbf{MOST related} (i.e., sentence A is closest in meaning to sentence B).

    \item the sentence pair that is the \textbf{LEAST related} (i.e., sentence A is farthest in meaning to sentence B).
\end{itemize}

\textbf{Sentence pairs can be related in many ways}. I.e., sentence A and sentence B can be related in different ways. The first pair of sentences in Table \ref{tab:eg_guidelines} are more related than the second one.
\begin{table*}[!ht]
\small
    \centering
    \begin{tabular}{c|l}
    \toprule
    \textbf{MOST Related Pair} & S1: The boy enjoyed reading under the lemon tree \\
                               & S2: There is a lemon tree next to the house
 \\ \midrule
    \textbf{LEAST related Pair} & S1: The boy enjoyed reading under the lemon tree \\
                                & S2: The boy was an excellent football player \\
\bottomrule
\end{tabular}
\caption{Example in the Guidelines. Examples of two pairs of sentences with different degrees of relatedness from \citet{abdalla-etal-2023-makes}.}
\label{tab:eg_guidelines}
\end{table*}
Often, sentence pairs that are more specific in what they share tend to be more related than sentence pairs that are only loosely about the same topic.

If a sentence has more than one interpretation, consider that meaning which is closest to the meaning of the other sentence in the pair. If both sentences have multiple meanings, then consider those meanings that are closest to each other.

\begin{table*}[!ht]
\small
    \centering
    \begin{tabular}{c|l}
    \toprule
    \textbf{Pair 1} & S1: The boy enjoyed reading under the lemon tree \\
                        &S2: I have a green hat \\
                            \midrule
    \textbf{Pair 2} & S1: The boy enjoyed reading under the lemon tree \\
    &S2: She was an excellent football player
     \\
    \bottomrule
    \end{tabular}
    \caption{Example in the Guidelines. Examples of two pairs of sentences that have similar degrees of relatedness where one can choose randomly the most vs. least related pairs (i.e., either Pair 1 or Pair 2).}
\end{table*}
If in the given set of four pairs, two (or more) sentence pairs are\textbf{ equally related} to each other and they are also the most related pairs, then select \textbf{either} one of them as the most related (i.e., \textbf{randomly}). Similarly, if two (or more) equally related pairs are also the least related pairs, then select either one of them as the least related. (See Table 2.)

\paragraph*{}
\textbf{You cannot select the same sentence pair for both categories.}

Try not to overthink the answer. Let your instinct guide you.

\subsection{Notes}
Sentence pairs can be related in many ways. Consider the entire meaning of the sentences before selecting the most related.  The sentences included in this task may contain foul language, though we have attempted to limit this.

\subsection{Examples (Q1)}
Which of the four sentence pairs in Table \ref{tab:q1} is MOST RELATED? Which pair is LEAST RELATED?
\begin{table*}[!t]
\small
    \centering
    \begin{tabular}{c|l}
    \toprule
    \textbf{Pair 1} & S1: My personal favorites from Narnia were the conversations between Aslan and Bree.\\
    &S2: This marks my progress through the Chronicles, picked up after reading The Narnia Code and Planet Narnia.
    \\
    \midrule
    \textbf{Pair 2} & S1: why won't she ask me out? \\
    &S2: and after all that you wont have to worry about getting a girl to like you. \\
    \midrule
    \textbf{Pair 3} & S1: A group of people are sitting on the grass outside of a rustic building. \\
    &S2: Group sitting on a grassy hill resting. \\ \midrule
    \textbf{Pair 4} & S1: If you change me back, I will feed each one of your snakes a large mouse!\\
    &S2: Offer people who join cash and coupons.
     \\
    \bottomrule
    \end{tabular}
    \caption{Q1 Example in the Guidelines.}
    \label{tab:q1}
\end{table*}
\subsubsection{A1}
\textbf{The most related pair} is Pair 3 because both sentences are talking about a group sitting/resting in grass. 

\textbf{The least related} is Pair 4 because Pair 4 sentences are completely unrelated, whereas the other pairs have some relatedness. 

\subsubsection{Note (A1)}
Pair 1 sentences are somewhat related, as they talk about Narnia/characters in that world (Aslan and Bree are characters in Narnia). However, the content of this sentence pair is not as related as Pair 3.

Pair 2 sentences are both talking about romantic relationships. 

\subsection{Examples (Q2)}
Which of the four sentence pairs in Table \ref{tab:q2} is MOST RELATED? Which pair is LEAST RELATED?
\begin{table*}[!t]
\small
    \centering
    \begin{tabular}{c|l}
    \toprule
\textbf{Pair 1} &
S1: That and a kids meal. \\
& S2: My two kids, ages 5 and 3! \\ \midrule
\textbf{Pair 2} &
S1: The spines , which may be up to 50 mm long , are modified hairs , mostly made of keratin .\\
&S2: The simplest shape is the long opening with a pointed arch known in England as the lancet . \\ \midrule
\textbf{Pair 3} &
S1: A woman wearing a white shirt and a red headband is sitting outside.\\
&S2: Two women stand outside a library. \\ \midrule
\textbf{Pair 4} &
S1: Ayodhya ,capital of King Rama is mentioned on the banks of Sarayu river . \\
&S2: Ramayana mentions that city of Ayodhya was situated on the bank of Sarayu river .\\ \bottomrule
    \end{tabular}
    \caption{Q2 Example in the Guidelines.} \label{tab:q2}
\end{table*}

\subsubsection{A2}
\textbf{The most related pair} is Pair 4. Both sentences are talking about the same city and mention that it is on the bank of river Sarayu.
\textbf{The least related pair} is Pair 2 because the sentences are completely unrelated. 
\subsubsection{Note (A2)}
Pair 3 sentences both refer to at least one woman outside. 

Pair 1 sentences refer to kids or kid-related things (making them slightly close in meaning).

\subsection{Examples (Q3)}
Which of the four sentence pairs in Table \ref{tab:q3} is MOST RELATED? Which pair is LEAST RELATED?

\begin{table*}[!t]
\small
    \centering
    \begin{tabular}{c|l}
    \toprule
\textbf{Pair 1} &
 S1: IBM has not shifted its focus from mainframes to compete with Windows \\
&S2: In 3 years, IBM has not been interested in the PC. \\ \midrule
\textbf{Pair 2} 
&S1: I wanted to see the scene where Quinn told the brotherhood he was in love with Blay. \\
&S2: I also would have liked to see the scene where Qhuinn asks Blay's dad for permission to propose to Blay. \\ \midrule
\textbf{Pair 3}
&S1: Jeremy desperately needs a stable home. \\
&S2: Furnishings were an angle bed, a stool, and a chamber pot on the dirt floor.\\ \midrule
\textbf{Pair 4}
&S1: That's difficult. They're both great \\
&S2: that's really hard they are both great!\\ \bottomrule
    \end{tabular}
    \caption{Q3 Example in the Guidelines.} \label{tab:q3}
\end{table*}
\subsubsection{A3}
\textbf{The most related pair} is Pair 4. Both sentences are paraphrases of each other. (Pair 1 and Pair 2 are quite related but not as exact paraphrases as Pair 4.)

\textbf{The least related pair} is Pair 3. Pair 3 sentences are somewhat related as they talk about house furnishings. However, they are still less related than all the other pairs.

\subsubsection{Note (A3)}
Pair 1 sentences both refer to IBM and their business strategy. We consider this to be more related than Pair 3 because it’s more specific in the details they share.

Pair 2 sentences talk about the same characters and their romantic situation.

\newpage

\section{Pre-trained models used}
We list down the various pre-trained HuggingFace models used in our experiments:
\begin{enumerate}
\itemsep-0.25em
    \item \href{https://huggingface.co/bert-base-multilingual-cased}{mBERT}
    \item \href{https://huggingface.co/FacebookAI/xlm-roberta-base}{XLMR}
    \item \href{https://huggingface.co/Davlan/afro-xlmr-large}{AfroXLMR}
     \item \href{https://huggingface.co/dccuchile/albert-base-spanish#:~:text=This%20is%20an%20ALBERT%20model,LR%3A%200.0008838834765}{ALBETO}
     \item \href{https://huggingface.co/uhhlt/am-roberta}{AmRoBERTa}
     \item \href{https://huggingface.co/UBC-NLP/ARBERTv2}{ARBERT}
      \item \href{https://huggingface.co/asafaya/bert-base-arabic}{\texttt{arb} BERT}
      \item \href{https://huggingface.co/dccuchile/bert-base-spanish-wwm-cased}{BETO}
      \item \href{https://huggingface.co/alger-ia/dziribert}{DziriBERT}
      \item \href{https://huggingface.co/ai4bharat/indic-bert}{Indic-BERT}
    \item \href{https://huggingface.co/UBC-NLP/MARBERT}{MARBERT}
    
    \item \href{https://huggingface.co/PlanTL-GOB-ES/roberta-base-bne}{RoBERTa-BNE}
    
    \item \href{https://huggingface.co/Davlan/xlm-roberta-base-finetuned-hausa}{HauRoBERTa}
    \item \href{https://huggingface.co/sentence-transformers/LaBSE}{LaBSE}
\end{enumerate}

\input{tables/sbert_baselines}
\input{tables/other}

\input{tables/example-gloss}

%% file: tables/data_sources.tex
\begin{table*}[h!]
\small
\centering
    \begin{tabular}{ll}
    \toprule[1.2pt]
        \textbf{Lang.} & \textbf{Datasets (\%)} \\ \midrule
        \texttt{afr} & Oscar (100\%).\\
        \texttt{amh} & News data (100\%).\\
        \texttt{arb} & Ted Talk subtitles on science, art, and society (96\%), Economy news data (4\%).\\
        \texttt{arq} & Conversational data (89\%), Youtube comments (11\%).\\
        \texttt{ary} & News data (100\%).\\
        \texttt{eng} & Wikipedia (29\%), ParaNMT (17\%), Formality (17\%), SNLI (8\%), Goodreads (22\%), STS (7\%).\\
        \texttt{esp} & MuchoChine (MC) (7\%), Spanish QC (4\%), PAWS-X (19.5\%), NLI-es (12\%), SICK-es (17\%), \\
        &STS (5\%), BSO (17.5\%), XL-Sum (18\%).\\
        \texttt{hau} & News data (100\%)\\
        \texttt{hin} & News data (100\%).\\
        \texttt{ind} & News data (82\%), Wikipedia/ROOTS (18\%).\\
        \texttt{kin} & News data (100\%).\\
        \texttt{mar} & News data (100\%).\\
        \texttt{tel} & News data (100\%).\\ \bottomrule
    \end{tabular}
\caption{The percentage of instances collected from different sources (datasets).}
\label{tab:data_collection}
\vspace*{-1mm}
\end{table*}

%% file: tables/sbert_baselines.tex
\begin{table}[h]
\centering
\begin{tabular}{ccc}
\toprule
\textbf{Language} & \textbf{Base} & \textbf{Finetuned} \\ \midrule
\texttt{afr} & 0.76  & 0.79  \\
\texttt{amh} & 0.79  & 0.85  \\
\texttt{arb} & 0.55  & 0.62  \\
\texttt{arq} & 0.40   & 0.60   \\
\texttt{ary} & 0.38  & 0.77  \\
\texttt{eng} & 0.82  & 0.83  \\
\texttt{esp} & 0.65  & 0.70   \\
\texttt{hau} & 0.48  & 0.69  \\
\texttt{hin} & 0.71  & 0.77  \\
\texttt{ind} & 0.53  & 0.50   \\
\texttt{kin} & 0.45  & 0.72  \\
\texttt{mar} & 0.82  & 0.88  \\
\texttt{tel} & 0.80   & 0.82  \\
 \bottomrule
\end{tabular}
\caption{Spearman correlation scores on LaBSE models with and without further fine-tuning on our training data (Base and fine-tuned, respectively).}
\label{tab:sbert_baselines}
\end{table}

%% file: tables/other.tex
\newpage
\begin{table}[h!]
\centering
\begin{tabular}{ccc}
\toprule
\textbf{Language} & \textbf{mBERT} & \textbf{XLMR} \\ \midrule
\texttt{afr} &	0.77 &	0.76 \\
\texttt{amh}	& 0.12 & 0.69 \\
\texttt{arb}	& 0.40	& 0.42 \\
\texttt{arq}	& 0.28 & 0.32 \\
\texttt{ary}	& 0.53 & 0.50 \\
\texttt{eng}	& 0.71	& 0.74 \\
\texttt{esp}	& 0.67	& 0.68 \\
\texttt{hau}	& 0.32	& 0.31 \\
\texttt{hin}	& 0.64	& 0.63 \\
\texttt{ind}	& 0.54	& 0.54 \\
\texttt{kin}	& 0.25	& 0.30 \\
\texttt{mar}	& 0.78	& 0.75 \\
\texttt{tel}	& 0.77	& 0.78 \\
\bottomrule
\end{tabular}
\caption {Spearman correlation of the BERTScore \cite{bert-score} with mBERT and XLMR on the different languages.}

\label{tab:other_baselines}
\end{table}

%% file: tables/example-gloss.tex
\begin{figure*}
    \centering
    \includegraphics[page=1, clip, trim={2.6cm 0 2.8cm 3cm}]
    {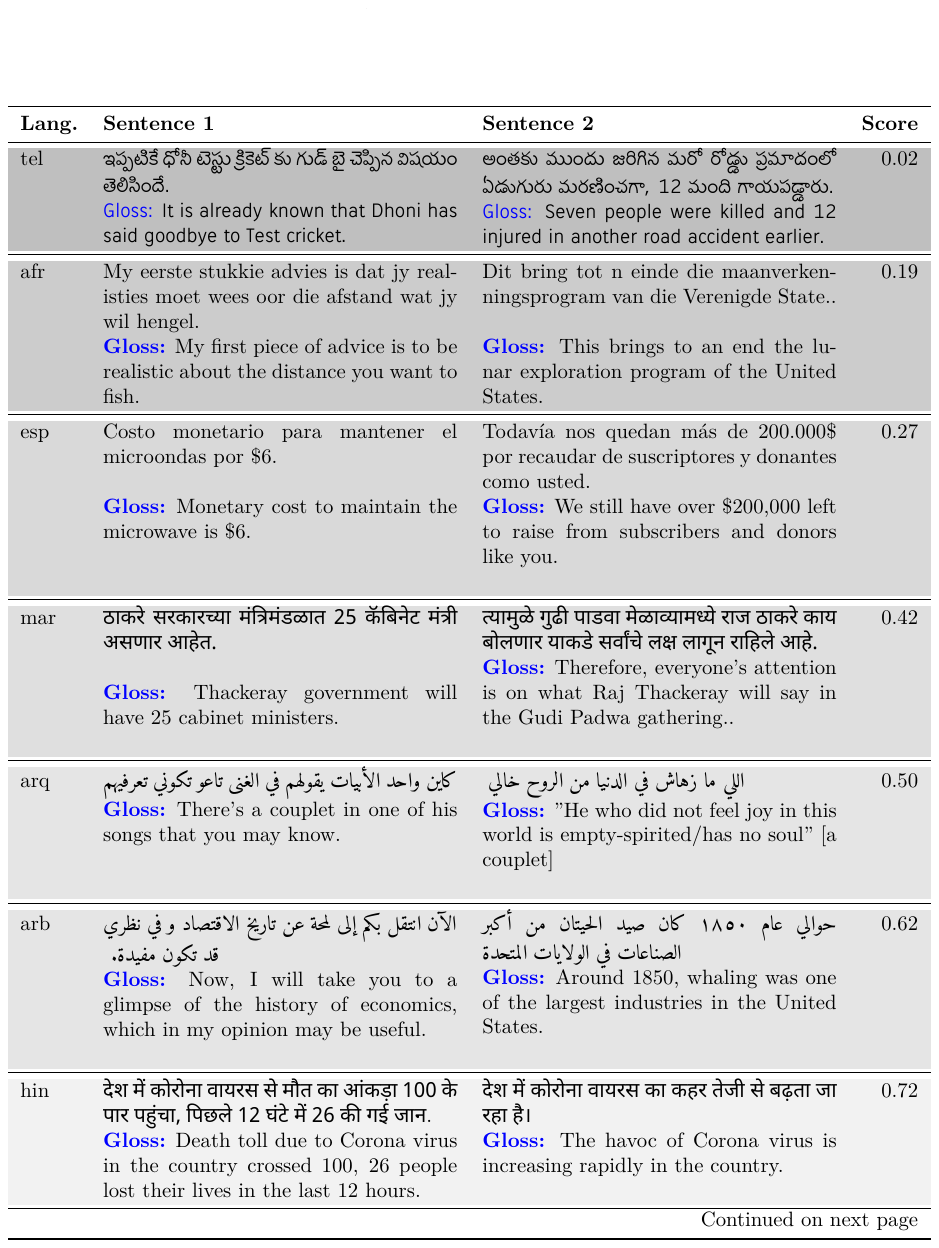}
\end{figure*}

\begin{figure*}
    \centering
    \includegraphics[page=2, clip, trim={2.6cm 0 2.8cm 2cm}]
    {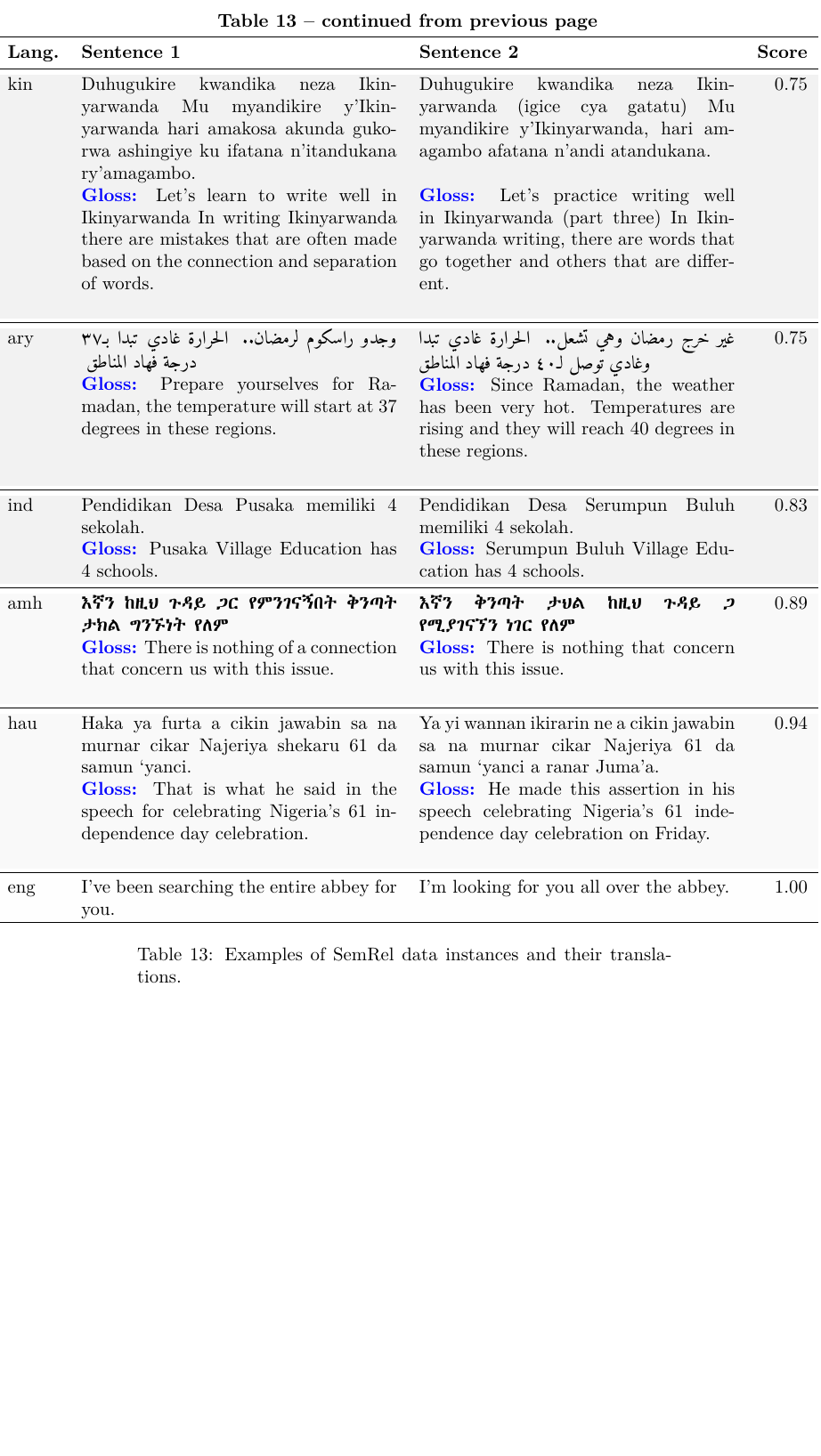}
\end{figure*}